%% file: main.tex
\documentclass[letterpaper]{article} 
\usepackage[submission]{aaai24}  
\usepackage{times}  
\usepackage{helvet}  
\usepackage{courier}  
\usepackage[hyphens]{url}  
\usepackage{graphicx} 
\usepackage{natbib}  
\usepackage{caption} 
\usepackage{amsmath}
\usepackage{algorithm}
\usepackage[noend]{algpseudocode}
\usepackage{arydshln}
\usepackage{amssymb}
\usepackage{enumitem, multirow, subfigure}
\usepackage{xcolor}
\usepackage{comment}
\usepackage{bm}
\usepackage{tabularray}
\usepackage{csquotes}
\usepackage{booktabs}

\usepackage{colortbl}
\definecolor{c_lowbest}{rgb}{0.4,0.9,0.9}
\definecolor{c_highbest}{rgb}{0.5,0.8,0.5}
\usepackage{arydshln}

\usepackage{newfloat}
\usepackage{listings}
\DeclareCaptionStyle{ruled}{labelfont=normalfont,labelsep=colon,strut=off} 
\floatstyle{ruled}
\newfloat{listing}{tb}{lst}{}
\floatname{listing}{Listing}
\title{ProNeRF: Learning Efficient Projection-Aware Ray Sampling for Fine-Grained Implicit Neural Radiance Fields}
\author{
    Juan Luis Gonzalez Bello\equalcontrib,
    Minh-Quan Viet Bui\equalcontrib,
    Munchurl Kim
}
\affiliations{
    Korea Advanced Institute of Science and Technology (KAIST), South Korea\\

}

\usepackage{bibentry}

\begin{document}

\maketitle

\begin{abstract}

  Recent advances in neural rendering have shown that, albeit slow, implicit compact models can learn a scene's geometries and view-dependent appearances from multiple views. To maintain such a small memory footprint but achieve faster inference times, recent works have adopted `sampler' networks that adaptively sample a small subset of points along each ray in the implicit neural radiance fields. Although these methods achieve up to a 10$\times$ reduction in rendering time, they still suffer from considerable quality degradation compared to the vanilla NeRF. In contrast, we propose ProNeRF, which provides an optimal trade-off between memory footprint (similar to NeRF), speed (faster than HyperReel), and quality (better than \textit{K}-Planes). ProNeRF is equipped with a novel projection-aware sampling (PAS) network together with a new training strategy for ray exploration and exploitation, allowing for efficient fine-grained particle sampling. Our ProNeRF yields state-of-the-art metrics, being 15-23$\times$ faster with 0.65dB higher PSNR than NeRF and yielding 0.95dB higher PSNR than the best published sampler-based method, HyperReel. Our exploration and exploitation training strategy allows ProNeRF to learn the full scenes' color and density distributions while also learning efficient ray sampling focused on the highest-density regions.
  We provide extensive experimental results that support the effectiveness of our method on the widely adopted forward-facing and 360 datasets, LLFF and Blender, respectively. 
\end{abstract}


\section{Introduction}
Neural radiance fields (NeRFs) \cite{nerf} have gained significant attention in the computer vision community due to their greater ability to compactly represent complex scenes' 3D geometries and view-dependent specularity, in comparison with other implicit representations \cite{deepview, siren}. The efficacy of NeRFs can be attributed to several key features such as: (i) the volumetric rendering technique \cite{volume_render}, which aggregates estimated RGB-density values along rendering rays, (ii) their implicit representation by a multi-layer perception (MLP) network that incorporates positional encoding \cite{nerf}, and (iii) their coarse-to-fine rendering strategy that enables \textit{dense} fine-grained ray sampling for high-quality rendering.

\begin{figure}
    \centering
    \includegraphics[width=0.8\linewidth]{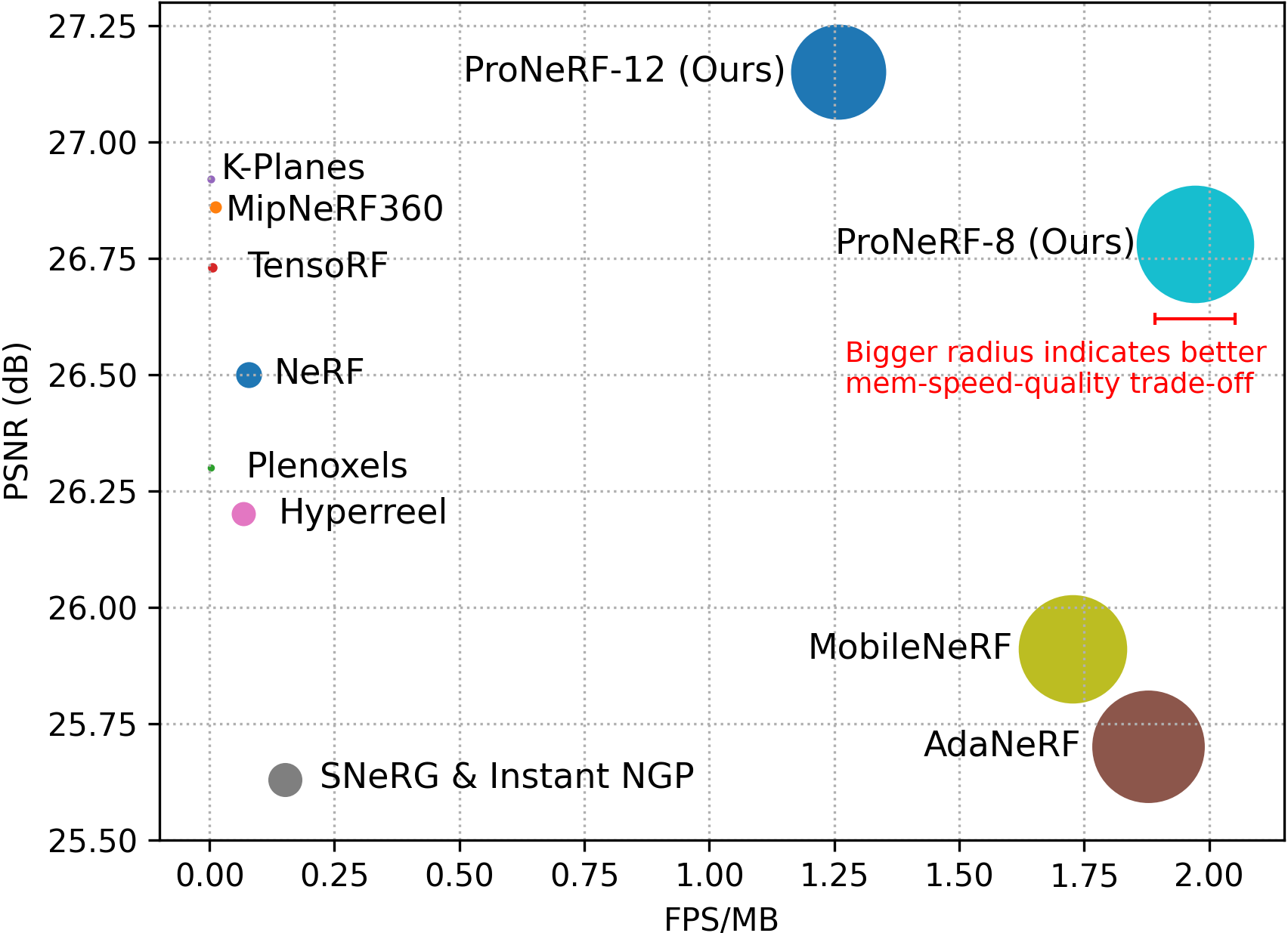}
    \caption{Performance trade-off of neural rendering (memory, speed, quality) on the LLFF dataset.}
\end{figure}

Although NeRFs offer a compact representation of 3D geometry and view-dependent effects, there is still significant room for improvement in rendering quality and inference times. To speed up the rendering times, recent trends have explored caching diffuse color estimation into an explicit voxel-based structure \cite{plenoctrees, snerg, fastnerf, efficient_nerf} or leveraging texture features stored in an explicit representation such as hash girds \cite{instantngp}, meshes \cite{mobilenerf}, or 3D Gaussians \cite{3Dgaussians}. While these methods achieve SOTA results on object-centric 360 datasets, they underperform for the forward-facing scene cases and require considerably larger memory footprints than NeRF. 

In a different line of work, the prior literature of \cite{donerf, terminerf, enerf, adanerf, hyperreel} has proposed training single-pass lightweight \enquote{sampler} networks, aimed to reduce the number of ray samples required for volumetric rendering. 
Although fast and memory compact, previous sampler-based methods often fall short in rendering quality compared to the computationally expensive vanilla NeRF.

In contrast, our proposed method with a Projection-Aware Sampling (PAS) network and an exploration-exploitation training strategy, denoted as \enquote{ProNeRF,} greatly reduces the inference times while simultaneously achieving superior image quality and more details than the current high-quality methods \cite{tensorf, kplanes_2023}. In conjunction with its small memory footprint (as small as NeRF), our ProNeRF yields the best performance profiling (memory, speed, quality) trade-off. Our main contributions are as follows\footnote{Visit our project website at \url{https://kaist-viclab.github.io/pronerf-site/}}:
\begin{itemize}[noitemsep, leftmargin=*]
\item \textbf{Faster rendering times}. Our ProNeRF leverages multi-view color-to-ray projections to yield a few precise 3D query points, allowing up to 23$\times$ faster inference times than vanilla NeRF under a similar memory footprint.

\item\textbf{Higher rendering quality}. Our proposed PAS and exploration-exploitation training strategy allow for \textit{sparse} fine-grained ray sampling in an end-to-end manner, yielding rendered images with improved quality metrics compared to the implicit baseline NeRF.

\item\textbf{Comprehensive experimental validation}. The robustness of ProNeRF is extensively evaluated on forward-facing and 360 object-centric multi-view datasets. Specifically, in the context of forward-facing scenes, ProNeRF establishes SOTA renders, outperforming implicit and explicit radiance fields, including NeRF, TensoRF, and \textit{K}-Planes with a considerably more optimal performance profile in terms of memory, speed, and quality.
\end{itemize}
\section{Related Work}
\label{sec:related_works}

The most relevant works concerning our proposed method focus on maintaining the compactness of implicit NeRFs while reducing the rendering times by \textbf{learning sampling networks} for efficient ray querying. 

Nevertheless, other works leverage data structures for \textbf{baking radiance fields}, that is, caching diffuse color and latent view-dependent features from a pre-trained NeRF to accelerate the rendering pipelines (as in SNeRG \cite{snerg}). Similarly, \citet{plenoctrees} proposed Plenoctrees to store spatial densities and spherical harmonics (SH) coefficients for fast rendering. Subsequently, to reduce the redundant computation in empty space, Plenoxels \cite{plenoxels} learns a sparse voxel grid of SH coefficients. On the other hand, Efficient-NeRF \cite{efficient_nerf} presents an innovative caching representation referred to as \enquote{NeRF-tree,} enhancing caching efficiency and rendering performance. However, these approaches require a pre-trained NeRF and a considerably larger memory footprint to store their corresponding scene representations.

Explicit data structures have also been used for storing latent textures in \textbf{explicit texture radiance fields} to speed up the training and inference times. Particularly, INGP \cite{instantngp} proposes quickly estimating the radiance values by interpolating latent features stored in multi-scaled hash grids. Drawing inspiration from tensorial decomposition, in TensoRF, \citet{tensorf} factorize the scene's radiance field into multiple low-rank latent tensor components. Following a similar decomposition principle, \citet{kplanes_2023} introduced \textit{K}-Planes for multi-plane decomposition of 3D scenes. Recently, MobileNeRF \cite{mobilenerf} and 3DGS \cite{3Dgaussians} concurrently propose merging the rasterization process with explicit meshes or 3D Gaussians for real-time rendering. Similar to the baked radiance fields, MobileNeRF and 3DGS demonstrate the capability to achieve incredibly rapid rendering, up to several hundred frames per second. However, they demand a considerably elevated memory footprint, which might be inappropriate in resource-constrained scenarios where real-time swapping of neural radiance fields is required, such as streaming, as discussed by \citet{adanerf}.

Inspired by the concept proposed in \cite{lightfield}, recent studies have also explored the learning of \textbf{neural light fields} which only require a single network evaluation for each casted ray. Light field networks such as LFNR \cite{nlf} and GPNR \cite{gpnr} presently exhibit optimal rendering performance across diverse novel view synthesis datasets. Nevertheless, they adopt expensive computational attention operations for aggregating multi-view projected features. Additionally, it's worth noting that similar to generalizable radiance fields (e.g., IBRNet \cite{ibrnet}, or NeuRay \cite{neuray}), LFNR and GPNR necessitate the storage of all training input images for epipolar feature projection, leading to increased memory requirements. Conversely, our method, ProNeRF, leverages color-to-ray projections while guaranteeing consistent memory footprints by robustly managing a small and fixed subset of reference views for rendering any novel view in the target scene. This eliminates the necessity for nearest-neighbor projection among all available training views in each novel scene. To balance computational cost and rendering quality for neural light fields, RSEN \cite{rsen} introduces a novel ray parameterization and space subdivision structure of the 3D scenes. On the other hand, R2L \cite{r2l} distills a compact neural light field with a pre-trained NeRF. Although R2L achieves better inference time and quality than RSEN, it necessitates the generation of numerous pseudo-images from a pre-trained NeRF to perform exhaustive training on dense pseudo-data. This process can extend over days of optimization.

In addition to IBRNet and NeuRay, other generalizable radiance fields have also been explored in \cite{pixelnerf, mine}, but are less relevant to our work. 

\textbf{Learning sampling networks}. In AutoInt, \citet{autoint} propose to train anti-derivative networks that describe the piece-wise color and density integrals of discrete ray segments whose distances are individually estimated by a sampler network. 
In DONeRF \cite{donerf} and TermiNeRF \cite{terminerf}, the coarse NeRF in vanilla NeRF is replaced with a sampling network that learns to predict the depth of objects' surfaces using either depth ground truth (GT) or dense depths from a pre-trained NeRF. The requirement of hard-to-obtain dense depths severely limits DONeRF and TermiNeRF for broader applications. ENeRF \cite{enerf} learns to estimate the depth distribution from multi-view images in an end-to-end manner. In particular, ENeRF adopts cost-volume aggregation and 3D CNNs to enhance geometry prediction.

Instead of predicting a continuous depth distribution, AdaNeRF \cite{adanerf} proposes a sampler network that maps rays to fixed and discretized distance probabilities. During test, only the samples with the highest probabilities are fed into the shader (NeRF) network for volumetric rendering. AdaNeRF is trained in a dense-to-sparse multi-stage manner without needing a pre-trained NeRF. The shader is first trained with computationally expensive dense sampling points, where sparsification is later introduced to prune insignificant samples, and then followed by simultaneous sampling and shading network fine-tuning. In MipNeRF360, \citet{mipnerf360} introduce online distillation to train the sampling network. Nevertheless, the sampler utilized in MipNeRF360 remains structured as a radiance field, necessitating a per-point forward pass. Consequently, incorporating this sampler does not yield substantial improvements in rendering latency. On the other hand, in the recent work of HyperReel, \citet{hyperreel} proposed a sampling network for learning the geometry primitives in grid-based rendering models such as TensoRF. HyperReel inherits the fast-training properties of TensoRF but also yields limited rendering quality with a considerably increased memory footprint compared to the vanilla NeRF.

Contrary to the existing literature, we present a sampler-based method, ProNeRF, that allows for fast neural rendering while substantially outperforming the implicit and explicit NeRFs quantitatively and qualitatively in reconstructing forward-facing captured scenes. The main components of ProNeRF are a novel PAS network and a new learning strategy that borrows from the reinforcement learning concepts of exploration and exploitation. Moreover, all the previous sampler-based methods require either pre-trained NeRFs (TermiNeRF), depth GTs (DoNeRF), complex dense-ray sampling and multi-stage training strategies (AdaNeRF), or large memory footprint (HyperReel). In contrast, our proposed method can more effectively learn the neural rendering in an end-to-end manner from sparse rays, even with shorter training cycles than NeRF.


\begin{figure*}
  \centering 
  \includegraphics[width=\textwidth]{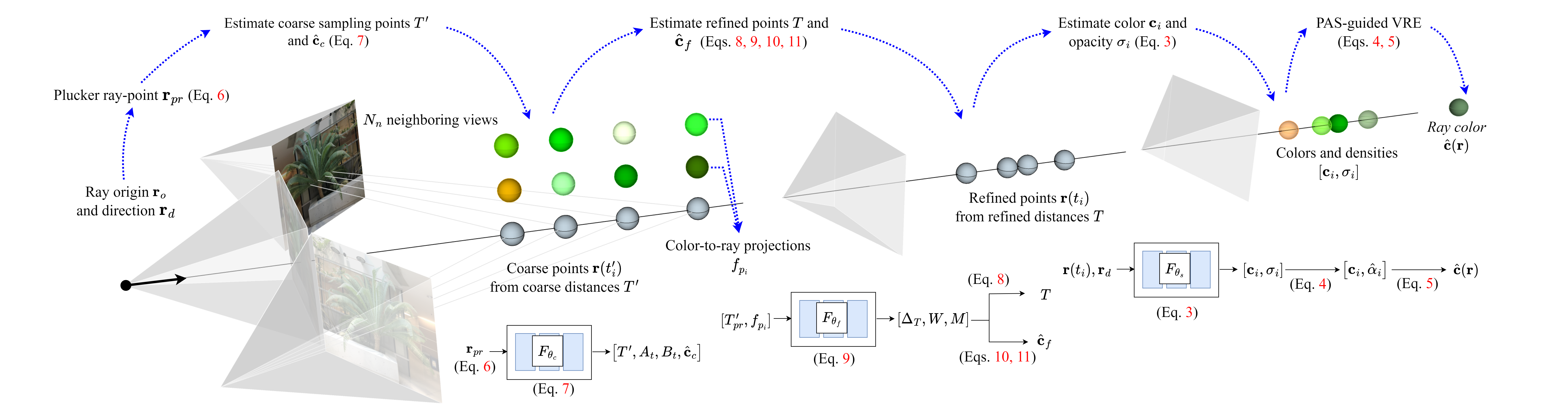}
  \caption{A conceptual illustration of our fast and high-quality projection-aware sampling of neural radiance fields (ProNeRF). The reference views are available during training and testing. The target view is drawn for illustrative purposes only.}
  \label{fig:full_pipeline}
\end{figure*}


\section{Proposed Method}
Fig. \ref{fig:full_pipeline} depicts a high-level overview of our ProNeRF, which is equipped with a projection-aware sampling (PAS) network and a shader network (a.k.a NeRF) for few-point volumetric rendering. ProNeRF performs PAS in a coarse-to-fine manner. First, for a given target ray, ProNeRF maps the ray direction and origin into coarse sampling points with the help of an MLP head ($F_{\theta_c}$). By tracing lines from these sampling points into the camera centers of the reference views \textit{in the training set}, ProNeRF performs a color-to-ray projection which is aggregated to the coarse sampling points and is processed in a second MLP head ($F_{\theta_f}$). $F_{\theta_f}$ then outputs the refined 3D points that are fed into the shading network ($F_{\theta_s}$) for the further volumetric rendering of the ray color $\hat{\bm{c}}$. See \textit{Section} \ref{sect:pas} for more details.

Training a ProNeRF as depicted in Fig. \ref{fig:full_pipeline} is not a trivial task, as the implicit shader needs to learn the \textit{full} color and density distributions in the scenes while the PAS network tries to predict ray points that focus on specific regions with the highest densities. Previous works, such as DoNERF, TermiNeRF, and AdaNeRF go around this problem at the expense of requiring depth GTs, pre-trained NeRF models, or expensive dense sampling. To overcome this issue, we propose an alternating learning strategy that borrows from reinforcement learning which (i) allows the shading network to \textit{explore} the scene's rays and learn the full scene distributions and (ii) leads the PAS network to \textit{exploit} the ray samples with the highest densities. See \textit{Section} \ref{sect: explore_exploit} for more details.

\subsection{PAS-Guided Volumetric Rendering}

Volumetric rendering synthesizes images by traversing the rays that originate in the target view camera center into a 3D volume of color and densities. As noted by \citet{nerf}, the continuous volumetric rendering equation (VRE) of a ray color $\bm{c}(\bm{r})$ can be efficiently approximated by alpha compositing, which is expressed as:
\begin{equation} \label{eq:vre}
\bm{\hat{\bm{c}}(\bm{r})} = {\textstyle\sum}^{N}_{i=1} \left( {\scriptstyle\prod}^{i-1}_{j=1} 1-\alpha_j \right) \alpha_i\bm{c}_i, 
\end{equation}
where $N$ is the total number of sampling points and $\alpha_i$ denotes the opacity at the $i^{th}$ sample in ray $\bm{r}$ as given by
\begin{equation} \label{eq:alpha}
\alpha_i = 1 - e^{-\sigma_i(t_{i+1} - t_i)}. 
\end{equation}
Here, $\sigma_i$ and $\bm{c}_i$ respectively indicate the density and colors at the 3D location given by $\bm{r}(t_i)$ for the $i^{th}$ sampling point on $\bm{r}$. A point on $\bm{r}$ in distance $t$ is $\bm{r}(t) = \bm{r}_o + \bm{r}_dt$ where $\bm{r_o}$ and $\bm{r_d}$ are the ray origin and direction, respectively. 

In NeRF \cite{nerf}, a large number of $N$ samples along the ray is considered to precisely approximate the original integral version of the VRE. In contrast, our objective is to perform high-quality volumetric rendering with a smaller number of samples $N_s << N$. Rendering a ray with a few samples in our ProNeRF can be possible by accurately sampling the 3D particles with the highest densities along the ray. Thanks to the PAS, our ProNeRF can yield a sparse set of accurate sampling distances, denoted as $T=\{t_1, t_2, ..., t_{N_s}\}$, by which the shading network $F_{\theta_s}$ is queried for each point corresponding to the ray distances in $T$ (along with $\bm{r}_d$) to obtain $\bm{c}_i$ and $\sigma_i$ as
\begin{equation} \label{eq:our_shader}
\left[ \bm{c}_i, \sigma_i\right] = F_{\theta_s}(\bm{r}(t_i), \bm{r}_d).
\end{equation}
Furthermore, similar to AdaNeRF, our ProNeRF adjusts the final sample opacities $\alpha_i$, which allows for fewer-sample rendering and back-propagation during training. However, unlike the AdaNeRF that re-scales the sample densities, we shift and scale the $\alpha$ values in our ProNeRF, yielding $\hat{\alpha}$:
\begin{equation} \label{eq:our_alpha}
\hat{\alpha}_i = a_i(1 - e^{-(\sigma_i + b_i)(t_{i+1} - t_i)}),
\end{equation}
where $a_i$ and $b_i$ are estimated by the PAS network as $A_t=\{a_1, a_2, ..., a_{N_s}\}$ and $B_t=\{b_1, b_2, ..., b_{N_s}\}$. We then render the final ray color in our \textit{PAS-guided VRE} according to
\begin{equation} \label{eq:vres}
\bm{\hat{c}}(\bm{r}) = {\textstyle\sum}^{N_s}_{i=1} \left( {\scriptstyle\prod}^{i-1}_{j=1} 1-\hat{\alpha}_j \right) \hat{\alpha}_i\bm{c}_i.
\end{equation}


\subsection{PAS: Projection-Aware Sampling}
\label{sect:pas}
Similar to previous sampler-based methods, our PAS network in the ProNeRF runs only once per ray, which is a very efficient operation during both training and testing. As depicted in Fig. \ref{fig:full_pipeline}, our ProNeRF employs two MLP heads that map rays into the optimal ray distances $T$ and the corresponding shift and scale in density values $A_t$ and $B_t$ required in the \textit{PAS-guided VRE}. 


The first step in the PAS of our ProNeRF is to map the ray's origin and direction ($\bm{r}_o$ and $\bm{r}_d$) into a representation that facilitates the mapping of training rays and interpolation of unseen rays.
Feeding the raw $\bm{r}_o$ and $\bm{r}_d$ into $F_{\theta_c}$ can mislead to overfitting, as there are a few ray origins in a given scene (as many as reference views). To tackle this problem, previous works have proposed to encode rays as 3D points (TermiNeRF) or as a Plücker coordinate which is the cross-product $\bm{r}_o \times \bm{r}_d$ (LightFields and HyperReel). Motivated by these works, we combine the Plücker and ray-point embedding into a `Plücker ray-point representation'. Including the specific points in the ray aids in making the input representation more discriminative, as it incorporates not only the ray origin but also the range of the ray, while the vanilla Plücker ray can only represent an infinitely long ray. The embedded ray $\bm{r}_{pr}$ is then given by
\begin{equation} \label{eq:plucker_ray}
\bm{r}_{pr} = [\bm{r}_d, \bm{r}_o + \bm{r}_d \odot \bm{t}_{nf}, (\bm{r}_o + \bm{r}_d \odot \bm{t}_{nf}) \times \bm{r}_d]
\end{equation}
where $\bm{t}_{nf}$ is a vector whose $N_{pr}$ elements are evenly spaced between the scene's near and far bounds ($t_n$ and $t_f$), $\odot$ is the Hadamard product, and $[\cdot, \cdot]$ is the concatenation operation. The ProNeRF processes the encoded ray $\bm{r}_{pr}$ via $F_{\theta_c}$ in the first stage of PAS to yield the coarse sampling distances $T' = \{t'_1, t'_2, ..., t'_{N_s}\}$ along $\bm{r}$. $F_{\theta_c}$ also predicts the shifts and scales in opacity values $A_t$ and $B_t$. Furthermore, inspired by light-fields, $F_{\theta_c}$ yields a light-field color output $\hat{\bm{c}}_{c}$ which is supervised to approximate the GT color $\bm{c}(\bm{r})$ to further regularize $F_{\theta_c}$ and improve the overall learning.
The multiple outputs of $F_{\theta_c}$ are then given by
\begin{equation} \label{eq:F1}
\left[T', A_t, B_t, \hat{\bm{c}}_{c}\right] = F_{\theta_c}(\bm{r}_{pr}).
\end{equation}


While the previous sampler-based methods attempt to sample radiance fields with a single network such as $F_{\theta_c}$, we propose a coarse-to-fine PAS in ProNeRF. In our ProNeRF, the second MLP head $F_{\theta_f}$ is fed with the coarse sampling points $\bm{r}(t'_i)$ and color-to-ray projections which are obtained by tracing lines between the estimated coarse 3D ray points and the camera centers of $N_n$ neighboring views from a pool of $N_t$ available images, as shown in Fig. \ref{fig:full_pipeline}. The pool of $N_t$ images in the training phase consists of all training images. However, it is worth noticing that only a significantly small number of $N_t$ views is needed for inference. The color-to-ray projections make ProNeRF projection-aware and enable $F_{\theta_f}$ to better understand the detailed geometry in the scenes as they contain not only image gradient information but also geometric information that can be implicitly learned for each point in space. That is, high-density points tend to contain similarly-valued multi-view color-to-ray projections. 

Although previous image-based rendering methods have proposed to \textit{directly} exploit projected reference-view-features onto the shading network, such as the works of \citet{nerf_attention} and \citet{nlf}, these approaches necessitate computationally expensive attention mechanisms and all training views storage for inference, hence increasing the inference latency and memory footprint. On the other hand, we propose to incorporate color-to-ray projections not for directly rendering the novel views but for \textit{fine-grained ray sampling of radiance fields}. As we learn to sample implicit NeRFs sparsely, our framework provides a superior trade-off between memory, speed, and quality.

The color-to-ray projections are concatenated with the Plücker-ray-point-encoded $\bm{r}'_{pr}$ of coarse ray distances $T'$, which is then fed into $F_{\theta_f}$, as shown in Fig. \ref{fig:full_pipeline}. In turn, $F_{\theta_f}$ improves $T'$ by yielding a set of inter-sampling refinement weights, denoted as $0 \leq \Delta_T \leq 1$. 
The refined ray distances $T$ are obtained by the linear interpolation between consecutive elements of the expanded set of coarse ray distances $\dot{T} =\{t_n, t'_1, t'_2, ..., t'_{N_s}, t_f\}$ from $T'$, as given by
\begin{equation} \label{eq:sample_refine}
T = \left\{\tfrac{1}{2} \left((\dot{T}_i + \dot{T}_{i+1}) + \Delta_{T_i}(\dot{T}_{i+2} - \dot{T}_{i})\right) \right\}^{N_s}_{i=1}.
\end{equation}
Our inter-sampling residual refinement aids in training stability by reusing and maintaining the order of the coarse samples $T'$. 
$\Delta_{T}$ is predicted by $F_{\theta_f}$ as given by
\begin{equation} \label{eq:ftheta2}
\left[\Delta_T, W, M \right] = F_{\theta_f}([\bm{r}'_{pr}, \bm{f}_{p_1}, \bm{f}_{p_2}, ..., \bm{f}_{p_{N_s}}]),
\end{equation}
where $\bm{f}_{p_i} = [{\bm{c}^1_{p_i}, \bm{c}^2_{p_i}, ..., \bm{c}^{N_n}_{p_i}}]$ and $\bm{c}^k_{p_i}$ is the $k^{th}$ color-to-ray projection from the $N_n$ views at 3D point $p_i = \bm{r}(t'_i)$. Note that $W$ and $M$ in Eq. (\ref{eq:ftheta2}) are the auxiliary outputs of softmax and sigmoid for network regularization, respectively.
In contrast with $F_{\theta_c}$, $F_{\theta_f}$ is projection-aware, thus $\hat{\bm{c}}_{f}$ is obtained by exploiting the color-to-ray projections in an \textit{approximated version of volumetric rendering} (AVR). In AVR,  $\bm{c}^k_{p_i}$ and $W\in \mathbb{R}^{N_s}$ are employed to approximate the VRE (Eq. \ref{eq:vre}). The terms $\left( {\scriptstyle\prod}^{i-1}_{j=1} 1-\alpha_j \right) \alpha_i$ in VRE are approximated by $W$ while $\bm{c}_i$ is approximated by projected color $\bm{c}^k_{p_i}$ for the $k^{th}$ view in $N_n$ neighbors. AVR then yields
\begin{equation} \label{eq:avre}
    \bm{c}^k_{avr} = {\textstyle\sum}^{N_s}_{i=1} W_i \bm{c}^k_{p_i},
\end{equation}
resulting in $N_n$ sub-light-field views. The final light-field output $\hat{\bm{c}}_{f}$ is aggregated by $M\in \mathbb{R}^{N_n}$ with $\bm{c}^k_{avr}$ as
\begin{equation} \label{eq:avre_combine}
    \hat{\bm{c}}_{f} = {\textstyle\sum}^{N_n}_{k=1} M_k\bm{c}^k_{avr}
\end{equation}




\begin{algorithm}
\small
\caption{Exploration and exploitation end2end training}
\begin{algorithmic}[1]
\Procedure{ProNeRF training}{}
\State Init Data, PAS, $F_{\theta_s}$, $Opt_{s}$, $Opt_{cfs}$
\For{$it=0$ to $7\times10^5$}
    \State Sample random ray $\bm{r}$
    \State $A_t$, $B_t$, $T$, $\hat{\bm{c}}_{c}$, $\hat{\bm{c}}_{f}$ $\leftarrow PAS(\bm{r})$
    \If {$2|it$ and $it$ $<$ 4$\times$$10^5$} \Comment{Exploration pass}
        \State $N^+_s \leftarrow RandInt(N_s, N)$
        \State $T^+ \leftarrow Sample(T, N^+_s)$
        \State $T^+ \leftarrow T^+ + noise$
        \State $\{\bm{c}_i, \sigma_i\}^{N^+_s}_{i=1} \leftarrow F_{\theta_s}(\bm{r}_o + \bm{r}_d \odot T^+)$
        \State $\hat{\bm{c}}(\bm{r}) \leftarrow VRE(\{\bm{c}_i, \sigma_i\}^{N^+_s}_{i=1}, T^+)$ (Eq. \ref{eq:vre})
        \State $loss \leftarrow |\hat{\bm{c}}(\bm{r})-\bm{c}(\bm{r})|_{2}$
        \State Back-propagate and update by $Opt_{s}$
    \Else \Comment{Exploitation pass}
        \State $\{\bm{c}_i, \sigma_i\}^{N_s}_{i=1} \leftarrow F_{\theta_s}(\bm{r}_o + \bm{r}_d \odot T)$
        \State $\hat{\bm{c}}(\bm{r}) \leftarrow VRE(\{\bm{c}_i, \sigma_i\}^{N_s}_{i=1}, A_t, B_t, T)$ (Eq. \ref{eq:vres})
        \State $loss \leftarrow |\hat{\bm{c}}(\bm{r})-\bm{c}(\bm{r})|_{2}$ 
        \If {$it$ $<$ 4$\times$$10^5$}
            \State $loss \leftarrow loss + |\hat{\bm{c}}_{c}-\bm{c}(\bm{r})|_{2} + |\hat{\bm{c}}_{f}-\bm{c}(\bm{r})|_{2}$
        \EndIf
        \State Back-propagate and update by $Opt_{cfs}$
    \EndIf
\EndFor
\EndProcedure
\end{algorithmic}
\label{alg:strategy}
\end{algorithm}

\subsection{Novel Exploration-Exploitation Training}
\label{sect: explore_exploit}
Our training strategy alternates between ray sampling \textit{exploration} and \textit{exploitation} as shown in Algorithm \ref{alg:strategy}. As noted in line(L)-2, we first initialize the dataset (composed of calibrated multi-views) by extracting the target rays and colors, followed by ProNeRF's networks' initialization. We implement two optimizers, one for exploration ($Opt_s$) and the other for exploitation ($Opt_{cfs}$). $Opt_s$ updates the weights in $F_{\theta_s}$, while $Opt_{cfs}$ updates all weights in $F_{\theta_c}, F_{\theta_f}, F_{\theta_s}$. The first step in a training cycle is to obtain the PAS outputs ($A_t$, $B_t$, $T$, $\hat{\bm{c}}_{c}$, $\hat{\bm{c}}_{f}$), as denoted in line 5 of Algorithm \ref{alg:strategy}. 

In the \textit{exploration} pass (Algorithm \ref{alg:strategy} L-7 to 13), $F_{\theta_s}$ learns the scene's full color and density distributions by randomly interpolating $N_s$ estimated $T$ distances into $N^+_s$ piece-wise evenly-spaced exploration sample distances $T^+$. For example, if the number of estimated ray distances is $N_s=8$ and the exploration samples are randomly set to $N^+_s=32$, the distance between each sample in $T$ will be evenly divided into four bins such that the sample count is 32. Moreover, we add Gaussian noise to $T^+$ as shown in of Algorithm \ref{alg:strategy} L-9, further allowing the $F_{\theta_s}$ to explore the scene's full color and density distributions. We then query $F_{\theta_s}$ for the $N^+_s$ exploration points to obtain $\bm{c}_i$ and $\sigma_i$ in the original VRE (Eq. \ref{eq:vre}). Finally, $F_{\theta_s}$ is updated in the \textit{exploration} pass.

In the \textit{exploitation} pass, described in Algorithm \ref{alg:strategy} L-15 to 20, we let the PAS and $F_{\theta_s}$ be greedy by only querying the samples corresponding to $T$ and using the PAS-guided VRE (Eq. \ref{eq:vres}). Additionally, we provide GT color supervision to the auxiliary PAS network light-field outputs $\hat{\bm{c}}_{c}$ and $\hat{\bm{c}}_{f}$ for the first 60\% of the training iterations. For the remaining 40\%, ProNeRF focuses on the \textit{exploitation} and disables the auxiliary loss as described by Algorithm \ref{alg:strategy} L-18 and 19.
Note that for rendering a ray color with a few points during \textit{exploitation} and testing, adjusting $\alpha_i$ in Eq. \ref{eq:our_alpha} is needed to compensate for the subsampled accumulated transmittance which is learned for the full ray distribution in the \textit{exploration} pass.

In summary, during \textit{exploration}, we approximate the VRE with Monte Carlo sampling, where a random number of samples, ranging from $N_s$ to $N$, are drawn around the estimated $T$. When training under \textit{exploitation}, we sparsely sample the target ray $\bm{r}$ given by $T$. Furthermore, we only update $F_{\theta_s}$ during the \textit{exploration} pass while using the original VRE (Eq. \ref{eq:vre}). However, in our \textit{exploitation} pass, we update all MLP heads while using the PAS-guided VRE (Eq. \ref{eq:vres}). See \textit{Section} \ref{sect:experiment} for more implementation details.

\subsection{Objective functions}
\label{sect:opt}
Similar to previous works, we guide ProNeRF to generate GT colors from the queried ray points with an $l_2$ penalty as
\begin{equation} \label{eq:c_l2}
l = \tfrac{1}{N_r}{\textstyle\sum}_{N_r} ||\hat{\bm{c}}(\bm{r})-\bm{c}(\bm{r})||_{2},
\end{equation}
which is averaged over the $N_r$ rays in a batch. In contrast with the previous sampler-based networks (TermiNeRF, AdaNeRF, DoNeRF, HyperReel), our ProNeRF predicts additional light-field outputs, which further regularize learning, and is trained with an auxiliary loss $l_a$, as given by 
\begin{equation} \label{eq:cs_l2}
l_{a} = \tfrac{1}{N_r}{\textstyle\sum}_{N_r} ||\hat{\bm{c}}_{c}(\bm{r})-\bm{c}(\bm{r})||_{2} + ||\hat{\bm{c}}_{f}(\bm{r})-\bm{c}(\bm{r})||_{2}.
\end{equation}
Our total objective loss is $l_T = l + \lambda l_{a}$, where $\lambda$ is 1 for 60\% of the training and then set to 0 afterward. 

\section{Experiments and Results}
\label{sect:experiment}
We provide extensive experimental results on the LLFF \cite{llff} and Blender \cite{nerf} datasets to show the effectiveness of our method in comparison with recent SOTA methods. Also, we present a comprehensive ablation study that supports our design choices and main contributions. More results are shown in \textit{Supplemental}.

We evaluate the rendering quality of our method by three widely used metrics: Peak Signal-to-Noise Ratio (PSNR), Structural Similarity (SSIM) \cite{ssim} and Learned Perceptual Image Patch Similarity (LPIPS) \cite{lpips}. When it comes to SSIM, there are two common implementations available, one from \textit{Tensorflow} \cite{tensorflow2015-whitepaper} (used in the reported metrics from NeRF, MobileNeRF, and IBRnet), and another from \textit{sci-kit image} \cite{scikit-image} (employed in ENeRF, RSeN, NLF). We denoted the metrics from \textit{Tensorflow} and \textit{scikit-image} as SSIM$_t$ and SSIM$_s$, respectively. Similarly, for LPIPS, we can choose between two backbone options, namely AlexNet \cite{alexnet} and VGG \cite{vggnet}. We present our SSIM and LPIPS results across all available choices to ensure a fair and comprehensive evaluation of our method's performance.

\subsection{Implementation Details}

We train our ProNeRF with PyTorch on an NVIDIA A100 GPU using the Adam optimizer with a batch of $N_r=4,096$ randomly sampled rays. The initial learning rate is set to $5\times10^{-4}$ and is exponentially decayed for 700K iterations. We used TensoRT on a single RTX 3090 GPU with model weights quantized to half-precision FP16 for testing. We set the point number in the Plücker ray-point encoding for our PAS network to 48. We set the maximum number of \textit{exploration} samples to $N = 64$. $F_{\theta_c}$ and $F_{\theta_f}$ consist of 6 fully-connected layers with 256 neurons followed by ELU non-linearities. Finally, we adopt the shading network introduced in DONeRF, which has 8 layers with 256 neurons.


\begin{figure*}
  \input{Figures/figure_qualitative_results}
  \caption{Qualitative comparisons for the LLFF \cite{llff} dataset. Zoom in for better visualization.}
  \label{fig:results_llff}
\end{figure*}
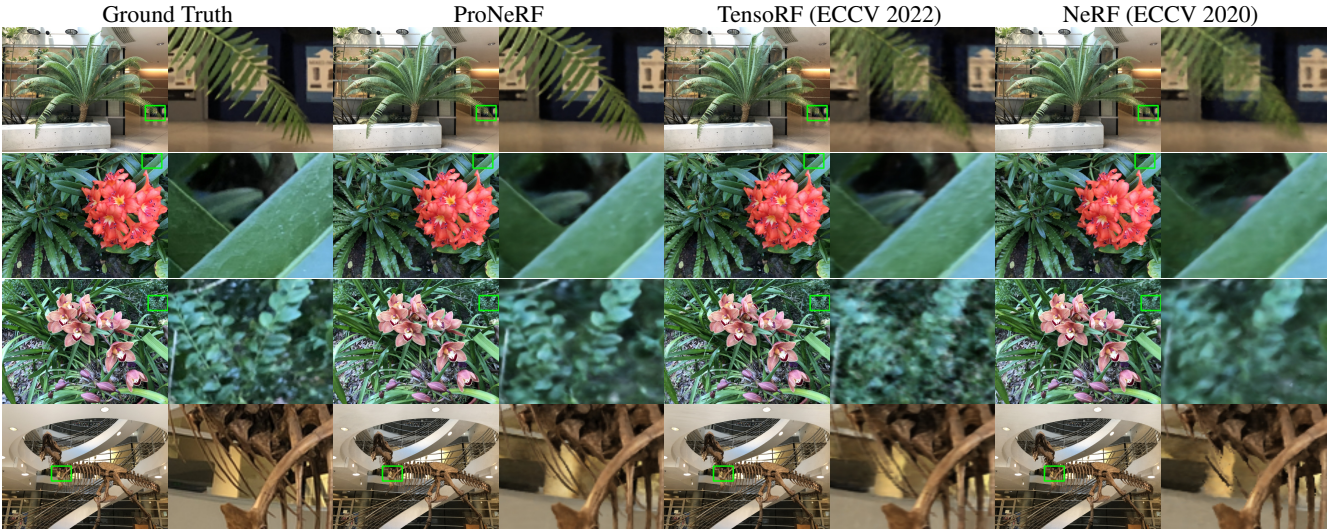

\subsection{Results}
\noindent
\textbf{Forward-Facing (LLFF).}
This dataset comprises 8 challenging real scenes with 20 to 64 front-facing handheld captured views. We conduct experiments on $756 \times 1008$ images to compare with previous methods, holding out every $8^{th}$ image for evaluation. We also provide the quantitative results on $378 \times 504$ images for a fair comparison to the methods evaluated on the lower resolution.

Our quantitative and qualitative results, respectively shown in Table \ref{tab:results_llff} and Fig. \ref{fig:results_llff}, demonstrate the superiority of our ProNeRF over the implicit NeRF and the previous explicit methods, e.g, TensoRF  and \textit{K}-Planes. Our model with 8 samples, ProNeRF-8, is the first sampler-based method that outperforms the vanilla NeRF by 0.28dB PSNR while being more than \textbf{20$\times$} faster. Furthermore, our ProNeRF-12 yields rendered images with \textbf{0.65dB} higher PSNR while being about \textbf{15$\times$} faster than vanilla NeRF. Our improvements are reflected in the superior visual quality of the rendered images, as shown in Fig. \ref{fig:results_llff}. On the lower resolution, ProNeRF-8 outperforms the second-best R2L by 0.28dB and the latest sampler-based HypeRreel by 0.58dB with faster rendering. 
In Table \ref{tab:results_llff}, compared to the explicit grid-based methods of INGP, Plenoxels and MobileNeRF, our ProNeRF shows a good trade-off between memory, speed, and quality.

We also present the quantitative results of the auxiliary PAS light field outputs in Table \ref{tab:results_llff}, denoted as PAS-8 $\bm{c}_f$ for both the regression (Reg) and AVR cases. We observed no difference in the final color output when Reg or AVR were used in ProNeRF-8. However, PAS-8 $\bm{c}_f$ (AVR) yields considerably better metrics than its Reg counterpart. 

Inspired by the higher FPS from PAS-8 $\bm{c}_f$ (AVR), we also explored \textit{pruning ProNeRF} by running the $F_{\theta_s}$ only for the \enquote{complex rays}. We achieve ProNeRF-8 prune by training a complementary MLP head $F_{\theta_m}$ which has the same complexity as $F_{\theta_c}$ and predicts the error between $\hat{\bm{c}}_f$ and $\hat{\bm{c}}$ outputs. When the error is low, we render the ray by PAS-8 $\bm{c}_f$ (AVR); otherwise, we subsequently run the shader network $F_{\theta_s}$. While \textit{pruning} requires an additional 3.3 MB in memory, the pruned ProNeRF-8 is 23\% faster than ProNeRF-8 with a small PSNR drop and negligible SSIM and LPIPS degradations, as shown in Table \ref{tab:results_llff}. Note that other previous sampler-based methods cannot be pruned similarly, as they do not incorporate the auxiliary light-filed output. Training \textit{pruning} is fast (5min). See more details in \textit{Supplemental}.


\noindent
\textbf{360 Blender.} This is an object-centric 360-captured synthetic dataset for which our ProNeRF-32 achieves a reasonably good performance of 31.92 dB PSNR, 3.2 FPS (after pruning) and 6.3 MB Mem. It should be also noted that the ProNeRF-32 outperforms NeRF, SNeRG, Plenoctree, and Plenoxels while still displaying a favorable performance profiling. See \textit{Supplemental} for detailed results.

\begin{table}[t]
    \scriptsize
    \centering
    \setlength{\tabcolsep}{1.0pt}
    \input{Tables/table_llff}
    \caption{Results on LLFF. Metrics are \colorbox{c_lowbest}{the lower the better} and \colorbox{c_highbest}{the higher the better}. (-) metrics are not provided in the original literature.}
    \label{tab:results_llff}
\end{table}


\begin{table}[t]
    \scriptsize
    \centering
    \setlength{\tabcolsep}{1pt}
    \input{Tables/table_ablation}
\input{Tables/table_ablation_noviews_infer}
    \caption{ProNeRF ablations on LLFF. (Left) Network designs on Fern. (Right) Ablation of \# of available ref. views.}
    \label{tab:ablation}
\end{table}

\subsection{Ablation Studies}
We ablate our ProNeRF on the LLFF's Fern scene in Table \ref{tab:ablation} (left). We first show that infusing exploration and exploitation into our \textbf{training strategy} is critical for high-quality neural rendering. As shown in the top section of Table \ref{tab:ablation} (left), exploration- or exploitation-only leads to sub-optimal results as neither the shading network is allowed to learn the full scene distributions nor the PAS network is made to focus on the regions with the highest densities.   

Next, we explore our \textbf{network design} by ablating each design choice. As noted in Table \ref{tab:ablation} (left), removing $\alpha$ scales ($A_t$) and shifts ($B_t$) severely impact the rendering quality. We also observed that the auxiliary loss ($l_a$) is critical to properly train our sampler since its removal causes almost 1dB drop in PSNR. 
The importance of our Plücker ray-point encoding is shown in Table \ref{tab:ablation} (left), having an impact of almost 0.5dB PSNR drop when disabled. Finally, we show that the color-to-ray projection in the PAS of our ProNeRF is the key feature for high-quality rendering. 

\begin{figure}[t]
    \centering
      \includegraphics[width=0.75\linewidth]{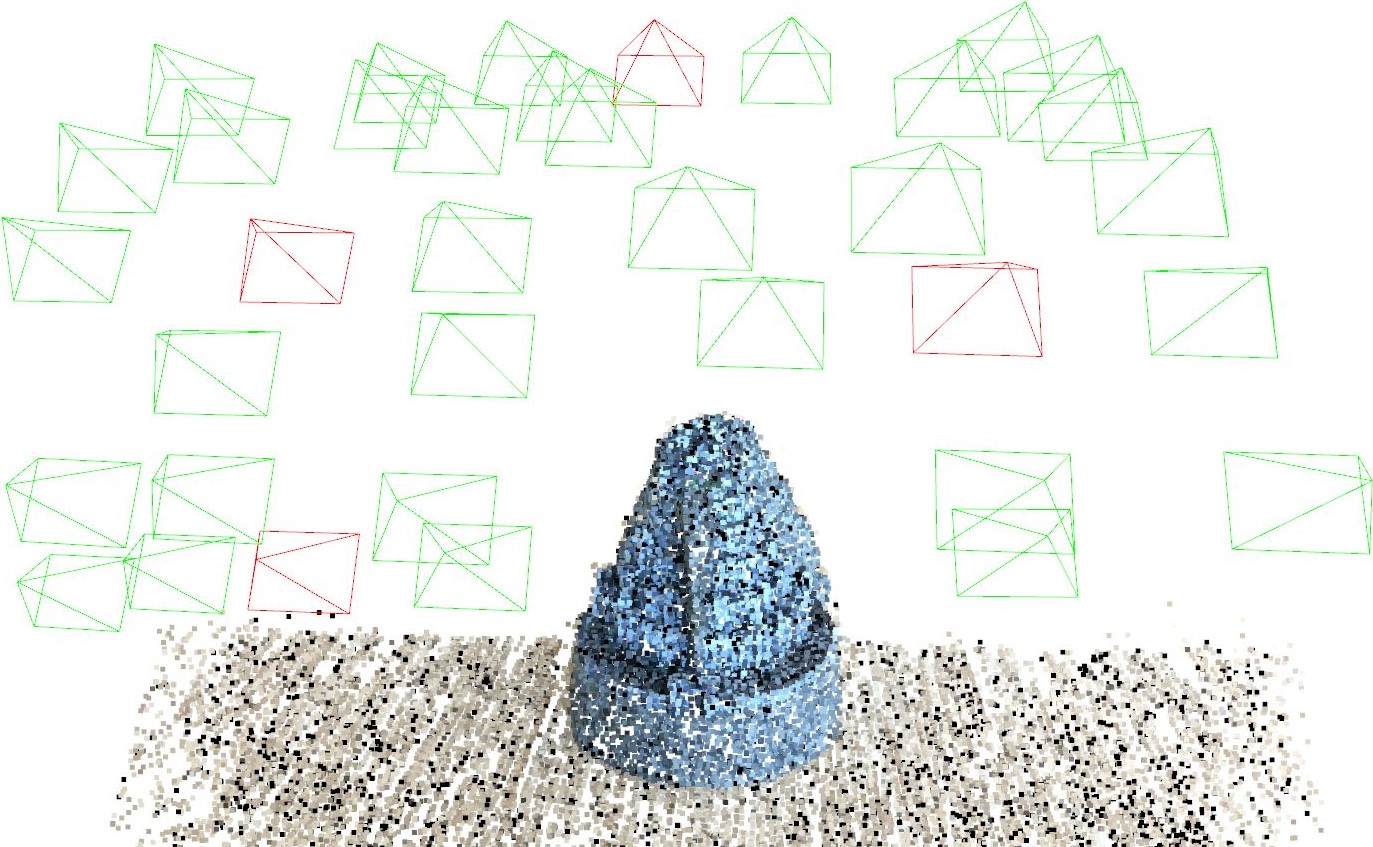}
    \caption{Cameras distribution on the LLFF's Fortress scene. \textcolor{green}{Green cameras} denote available training views. \textcolor{red}{Red cameras} denote selected and fixed subset of $N_t$ frames for projection.}
    \label{fig:ref_views}
\end{figure}

\textbf{Memory footprint consistency.}
This experiment proves ProNeRF yields a consistent usage of memory footprint. 
As mentioned in \textit{Section} \ref{sec:related_works}, light-fields and image-based rendering methods, which rely on multi-view color projections, typically require large storage for all available training views for rendering a novel view. This is because they utilize the nearest reference views to the target pose from the entire pool of available images. 
In contrast, our ProNeRF takes a distinct approach by consistently selecting a \textbf{fixed} subset of $N_t$ reference views when rendering any novel viewpoint in the inference stage. This is possible because (i) we randomly select any $N_n$ neighboring views (from the entire training pool) \textit{during training}; and (ii) our final rendered color is obtained by sparsely querying a radiance field, not by directly processing projected features/colors.
As a result, our framework yields a consistent memory footprint for storing reference views, which is advantageous for efficient hardware design. To select the $N_t$ views, we leverage the sparse point cloud reconstructed from COLMAP and a greedy algorithm to identify the optimal combination of potential frames. As shown in Fig. \ref{fig:ref_views}, the $N_t$ views become a subset across all available training images that comprehensively cover the target scene (see details in \textit{Supplemental}). As shown in Table \ref{tab:ablation} (right), we set the number of neighbors in PAS to $N_n=4$ and adjust $N_t$ to 4, 8, 12, and all training views (32.75). Please note our ProNeRF's rendering quality remains stable while modulating $N_t$, attesting to the stability and robustness of our approach across varying configurations.


\subsection{Limitations}
While not technically constrained to forward-facing scenes (such as NeX) and yielding better metrics than vanilla NeRF and several other works, our method is behind grid-based explicit models such as INGP for the Blender dataset. The methods like INGP contain data structures that better accommodate these kinds of scenes. Our method requires more samples for this data type, evidencing that our method is more efficient and shines on forward-facing datasets.


\section{Conclusions}
Our ProNeRF, a sampler-based neural rendering method, significantly outperforms the vanilla NeRF quantitatively and qualitatively for the first time. It also outperforms the existing explicit voxel/grid-based methods by large margins while preserving a small memory footprint and fast inference. Furthermore, we showed that our exploration and exploitation training is crucial for learning high-quality rendering. Future research might extend our ProNeRF for dynamic-scenes and cross-scene generalization.

\section*{Acknowledgements}
This work was supported by IITP grant funded by the Korea government (MSIT) (No. RS2022-00144444, Deep Learning Based Visual Representational Learning and Rendering of Static and Dynamic Scenes).

\bibliography{aaai24}

\end{document}

%% file: Figures/figure_qualitative_results.tex
\begin{tblr}{
  stretch = 0,
  colsep  = 0pt,
  colspec = {Q[h,c]Q[h,c]Q[h,c]Q[h,c]},
  rowsep = 0.3pt,
}
  \small{Ground Truth} & \small{ProNeRF} & \small{TensoRF (ECCV 2022)} & \small{NeRF (ECCV 2020)}\\
  \includegraphics[width=44mm]{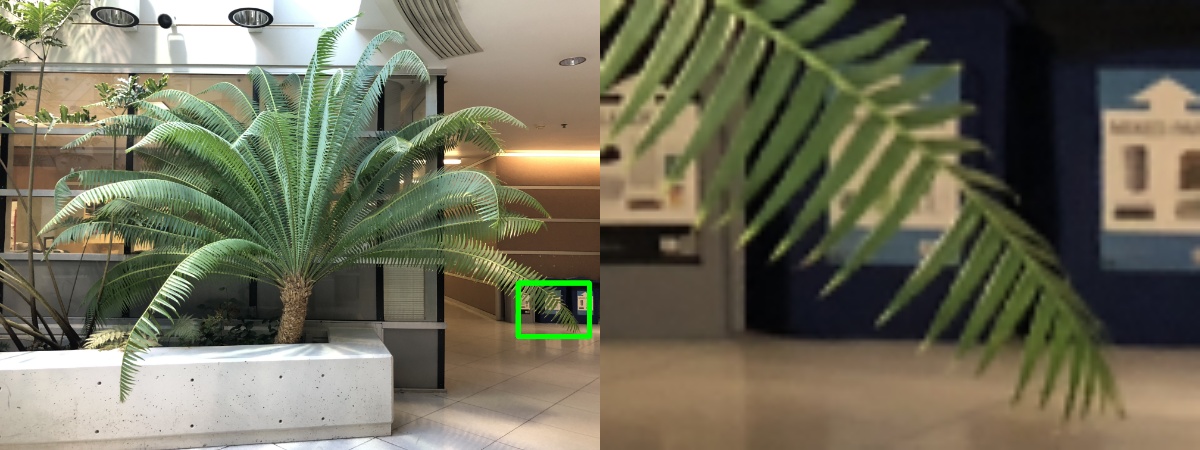} &\includegraphics[width=44mm]{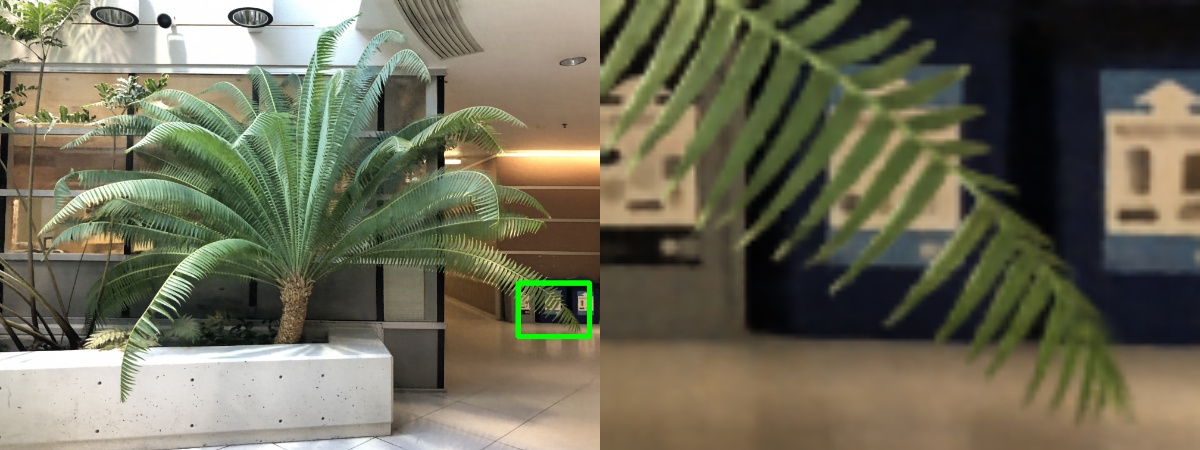}&\includegraphics[width=44mm]{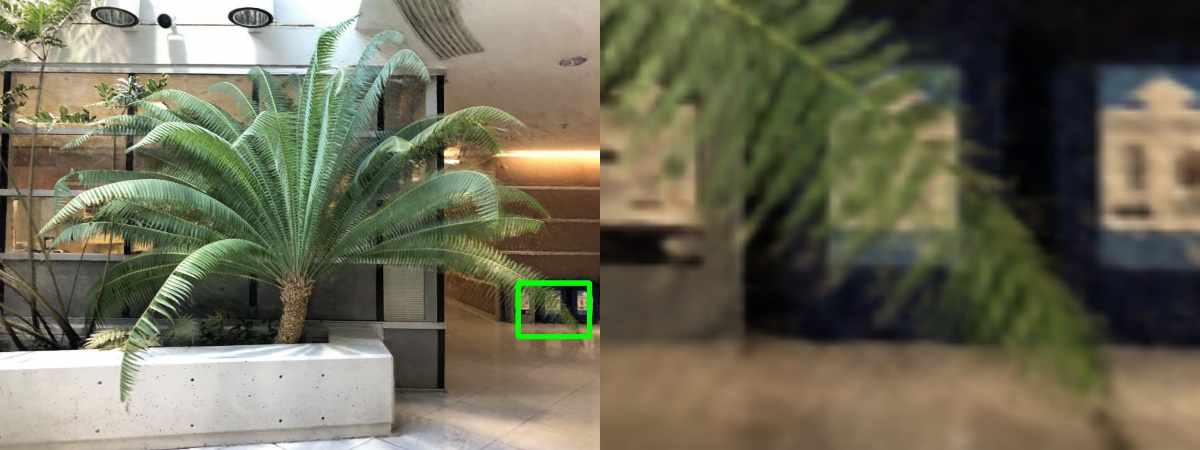} &\includegraphics[width=44mm]{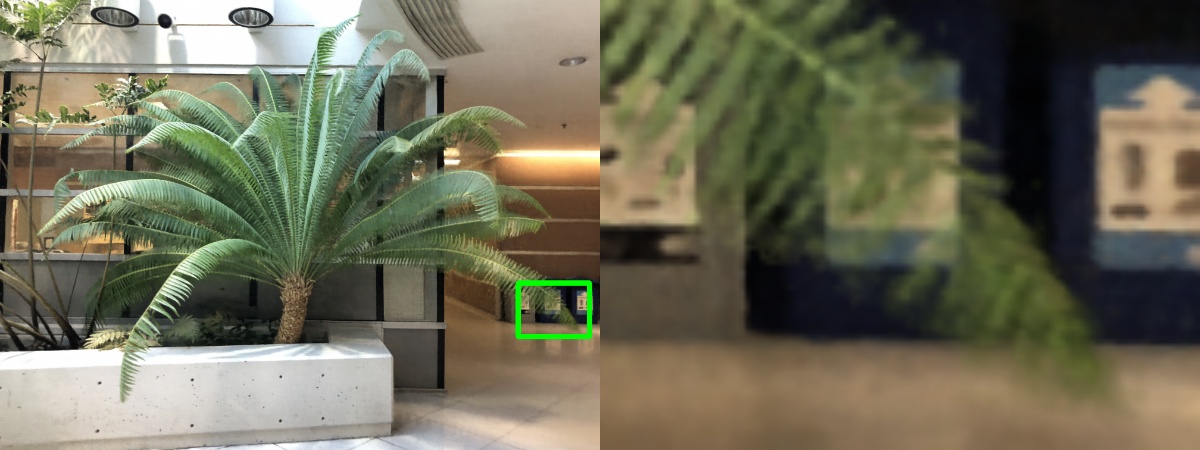}\\
  \includegraphics[width=44mm]{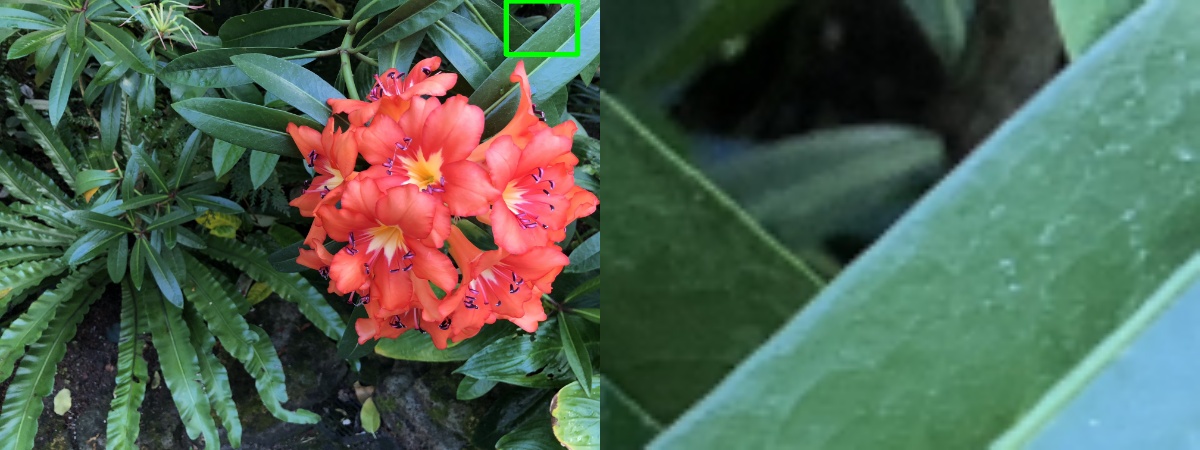} &\includegraphics[width=44mm]{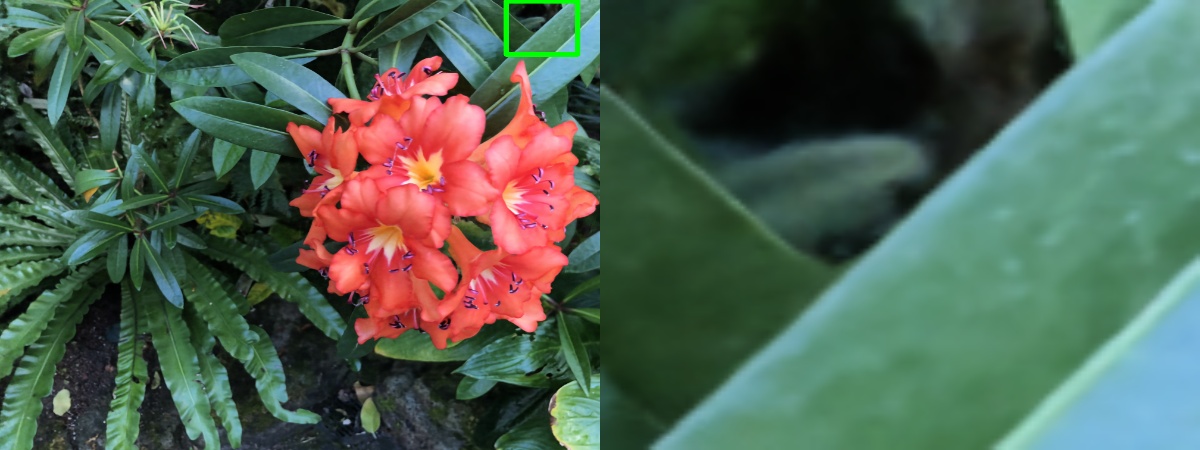}&\includegraphics[width=44mm]{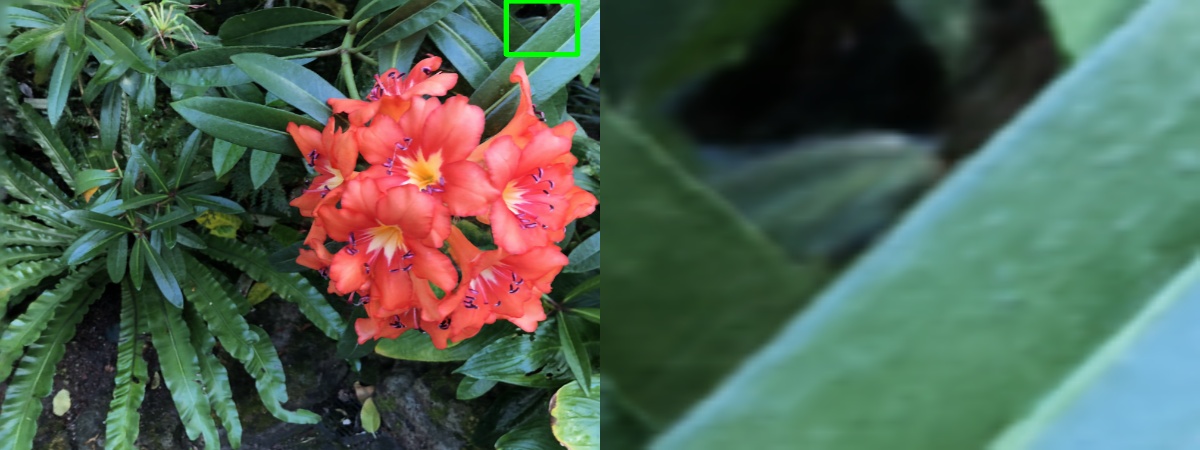} &\includegraphics[width=44mm]{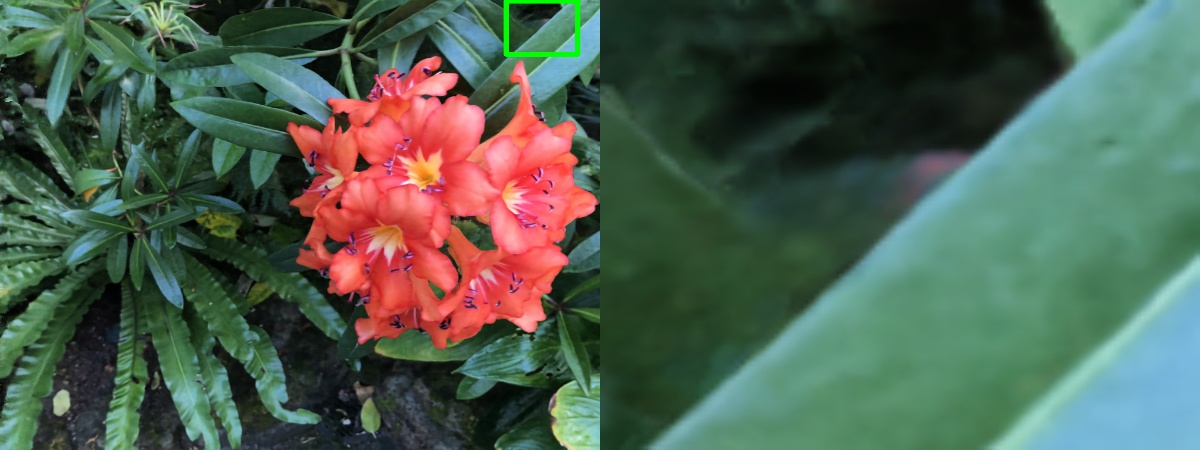}\\
  \includegraphics[width=44mm]{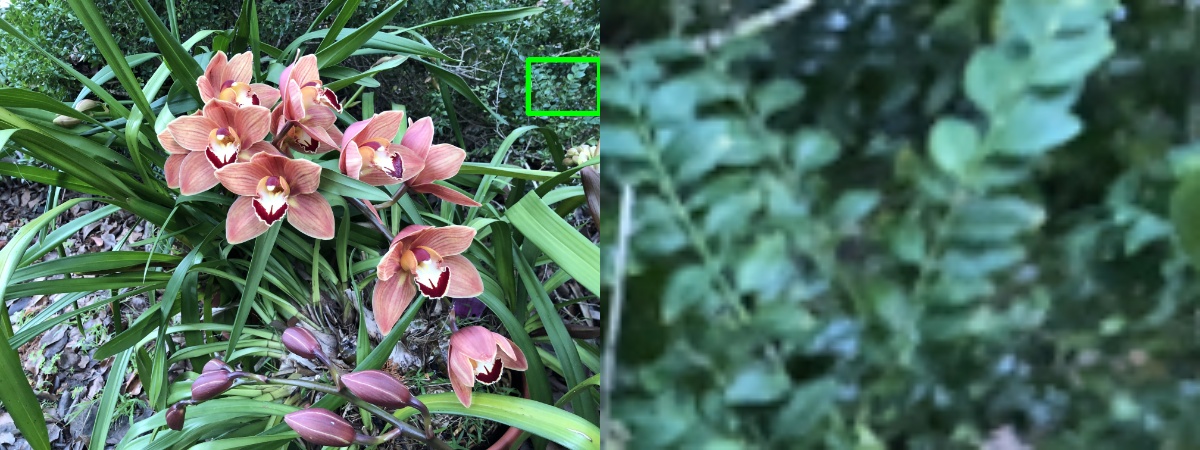} &\includegraphics[width=44mm]{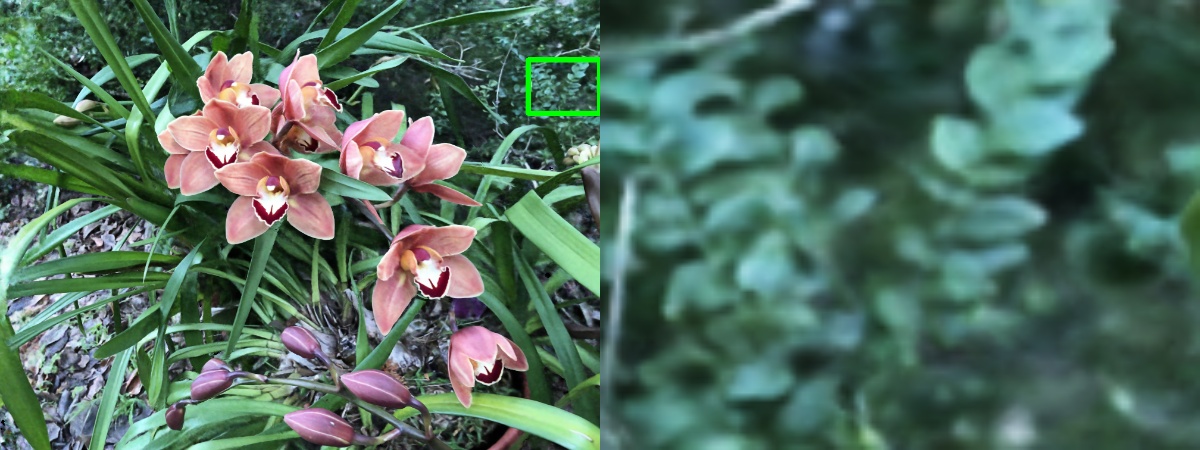}&\includegraphics[width=44mm]{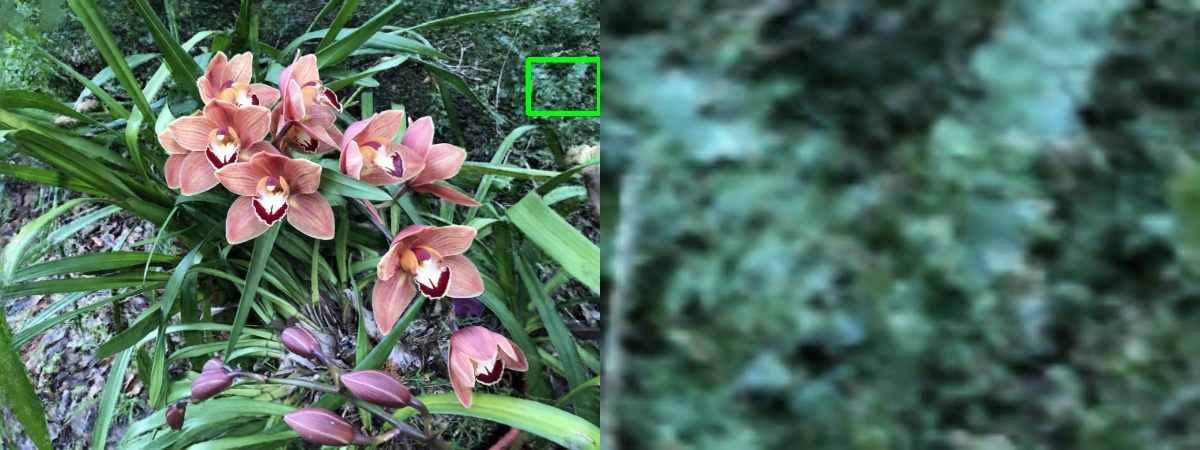} &\includegraphics[width=44mm]{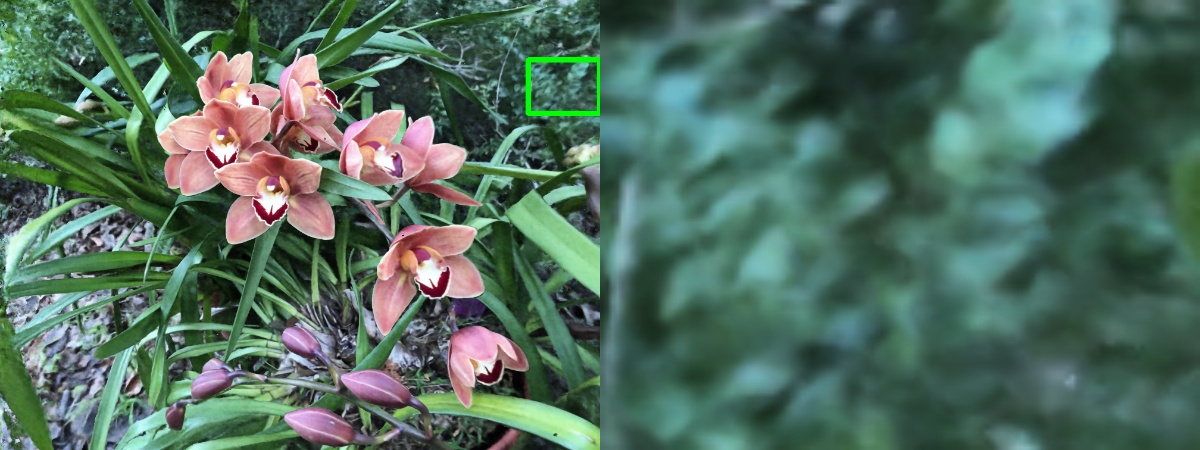}\\
  \includegraphics[width=44mm]{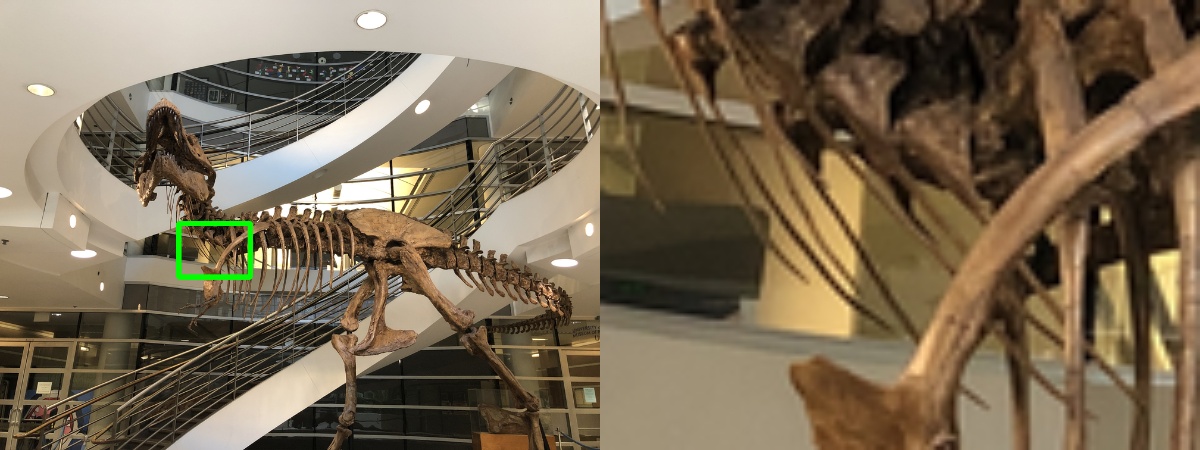} &\includegraphics[width=44mm]{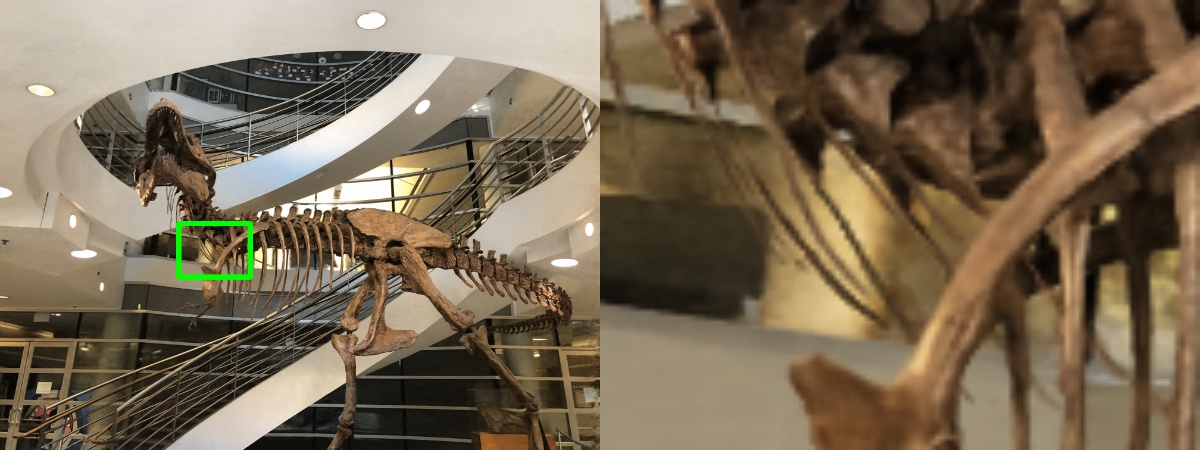}&\includegraphics[width=44mm]{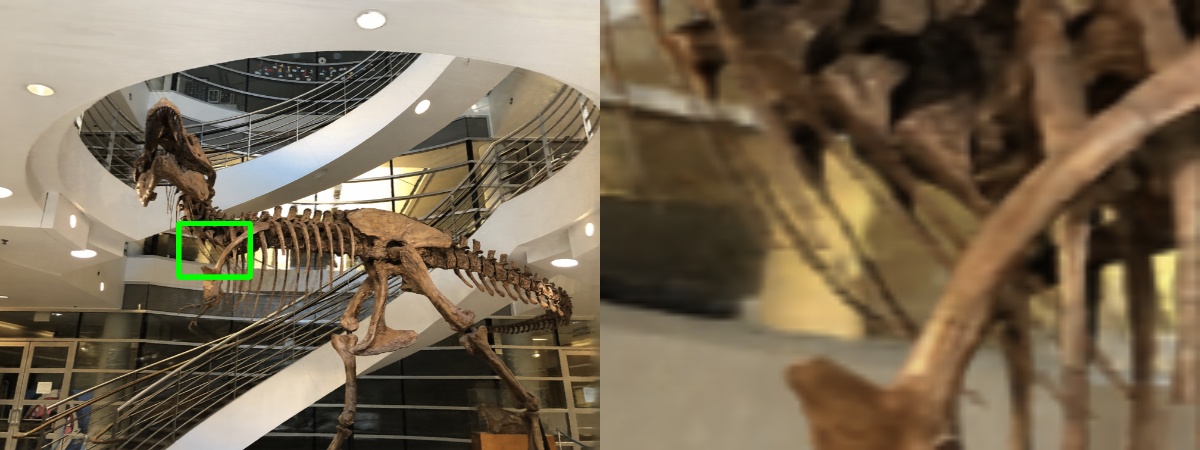} &\includegraphics[width=44mm]{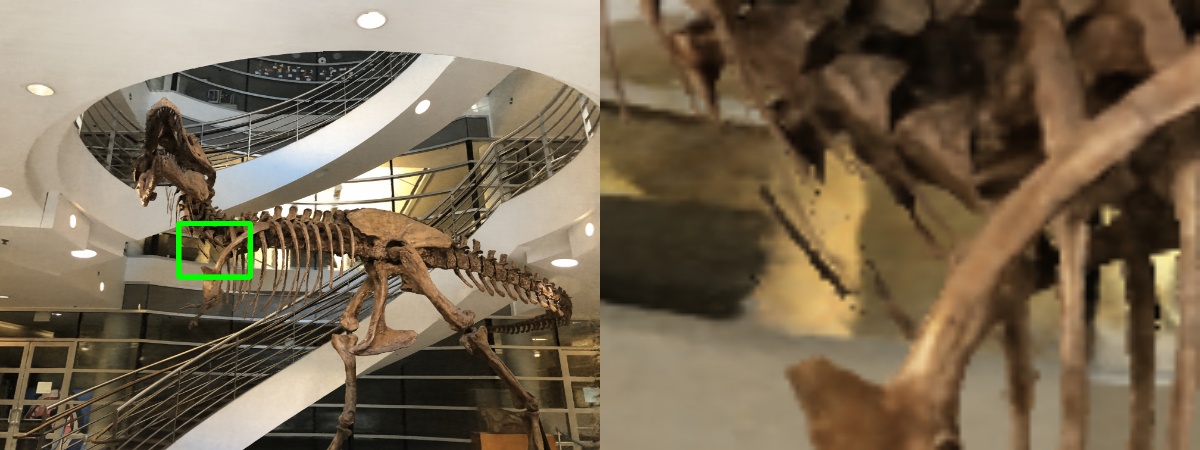}\\
\end{tblr}

%% file: Tables/table_llff.tex
\begin{tabular}{clccccc}
\hline
Res. & Methods  & PSNR\cellcolor{c_highbest} & SSIM$_{t/s}$\cellcolor{c_highbest} & LPIPS$_{vgg/alex}$\cellcolor{c_lowbest} & FPS\cellcolor{c_highbest} & Mem(MB)\cellcolor{c_lowbest} \\ 
\hline

 & NeRF (ECCV20) & 26.50 & 0.811 / - & 0.250 / - & 0.3  & 3.8 \\
&INGP (SIGGRAPH22)   & 25.60 & 0.758 / - & 0.267 / - & 7.3 & 64.0 \\
\multirow{9}{*}{$756$}&Plenoxels (CVPR22) & 26.30 &  0.839 / - & 0.210 / - & 9.1 & 3629.8 \\
\multirow{9}{*}{$ \times $}& MipNeRF360 (CVPR22) & 26.86 & \textbf{0.858} / - & - / 0.128 &  0.1 &  8.2 \\
\multirow{9}{*}{$1008$}&TensoRF (ECCV22)   & 26.73 & 0.839 / - & 0.204 / 0.124 & 1.1 & 179.7 \\
&K-Planes (CVPR23)  & 26.92 &  {0.847} / - & 0.182 / - & 0.7 & 214 \\
&SNeRG (ICCV21)  & 25.63 & 0.818 / - & 0.183 / - & 50.7 & 337.3 \\
&ENeRF (SIGGRAPHA22) & 24.89 & - / 0.865 & \textbf{0.159} / - & 8.9 &  10.7\\
&AdaNeRF (ECCV22) & 25.70 & - / - & - / - & 7.7 &  4.1 \\
&Hyperreel (CVPR23) & 26.20 & - / - & - / - & 4.0 & 58.8 \\
&MobileNeRF (CVPR23) & 25.91 & 0.825 / - & 0.183 / - & \textbf{348} & 201.5 \\

\cdashline{2-7}

&PAS-8 $\bm{c}_{f}$ (Reg) (Ours) & 24.86 & 0.787 / 0.855 & 0.236 / 0.150 & 29.4 &  \textbf{2.7}  \\
&PAS-8 $\bm{c}_{f}$ (AVR) (Ours) & 25.15 & 0.793 / 0.860 & 0.234 / 0.146 & 25.6 &  5.0  \\
&ProNeRF-8 Prune (Ours)& 26.54 & 0.825 / 0.883 & 0.219 / 0.120 & 8.5 &  6.8 \\
&ProNeRF-8 (Ours)  & 26.78 & 0.825 / 0.884 & 0.228 / 0.119 & 6.9 &  3.5 \\
&ProNeRF-12 (Ours) & \textbf{27.15} & 0.838 / \textbf{0.894} & 0.217 / \textbf{0.109} & 4.4 &  3.5 \\
\hline

 & FastNeRF (ICCV21) & 26.04 & - / 0.856 & - / 0.085 & \textbf{700}  & 4100 \\
\multirow{3}{*}{$378$} & EfficientNeRF (CVPR22)  & 27.39 & - / 0.912 & - / 0.082 & 219 & 2800 \\
\multirow{3}{*}{$ \times $}&RSEN (CVPR22) & 27.45 & - / 0.905 & - / 0.060 & 0.34 & 5.4 \\
\multirow{3}{*}{$504$}&R2L (ECCV22) & 27.79 &  - / - & - / 0.097 & 5.6 & 22.6 \\
&Hyperreel (CVPR23) & 27.50 & - / - & - / - & 4.0 & 58.8 \\

\cdashline{2-7}

&ProNeRF-8 (Ours)& 28.08 & 0.879 / 0.916 & 0.129 / 0.060 & 6.9 &  \textbf{3.5} \\
&ProNeRF-12 (Ours) & \textbf{28.33} & \textbf{0.885 / 0.920} & 0.129 / \textbf{0.058} & 4.4 & \textbf{3.5} \\

\hline

\end{tabular}

%% file: Tables/table_ablation.tex
\begin{tabular}{lcccc}
\hline
Methods & PSNR\cellcolor{c_highbest} & SSIM\cellcolor{c_highbest} & LPIPS\cellcolor{c_lowbest} \\ 
\hline
No exploration pass  & 24.00 & 0.754 & 0.299 \\
No exploitation pass  & 24.31 & 0.779 & 0.278 \\
\hline
No $\sigma$ shift (no $B_t$) & 24.2 & 0.773 & 0.264 \\
No aux. loss (no $l_a$) & 24.26 & 0.766 & 0.296 \\
No $\hat{\alpha}$ (no $A_t, B_t$)   & 24.69 & 0.785 & 0.260 \\
No Plücker ray-point & 24.72 & 0.782 & 0.257 \\
No color-to-ray proj  & 24.83 & 0.789 & 0.245 \\
\hline
ProNeRF-12 $N_n$=4 & \textbf{25.17} & \textbf{0.809} & \textbf{0.244} \\
\hline

\end{tabular}

%% file: Tables/table_ablation_noviews_infer.tex
\begin{tabular}{cccccc}
\hline
Avg $N_t$ & PSNR\cellcolor{c_highbest} & SSIM\cellcolor{c_highbest} & LPIPS\cellcolor{c_lowbest} & Mem(MB)\cellcolor{c_lowbest}\\ 
\hline
 4.00 &  27.15 & \textbf{0.838} & 0.217 & \textbf{3.5}\\
 8.00 &  \textbf{27.16} & \textbf{0.838} & \textbf{0.216} & 4.2\\
 12.00  & 27.15 & 0.837 & 0.217 & 4.9\\
\hline
 32.75 & {27.15} & \textbf{0.838} & \textbf{0.216} & 8.4\\
\hline

\end{tabular}
